# Temporal Tree of Thought: Reasoning-Guided Visual Cue Search for Long-Video Understanding

**Ziling Huang**
National Institute of Informatics
huangziling@nii.ac.jp

**Shin'ichi Satoh**
National Institute of Informatics
satoh@nii.ac.jp

## Abstract

Long-video understanding remains challenging for Multimodal Large Language Models (MLLMs) due to limited context length. Uniform sampling may miss crucial moments, while agent-based frame video understanding methods often evaluate frames independently, overlooking the temporal organization of videos. Ideally, evidence selection should mimic how humans answer questions about long videos: first locating the relevant segment from the global context, then zooming into local objects, and details. We propose Temporal Tree of Thought ($T^3$), a training-free framework for adaptive coarse-to-fine long-video understanding. $T^3$ constructs a question-agnostic hierarchical temporal tree via recursive temporally constrained clustering, where each node represents a contiguous segment with an informative key frame. During inference, $T^3$ performs an answer–retrieve–explore loop: it reasons over coarse representative frames, generates a search statement when evidence is insufficient, and expands relevant branches for finer-grained evidence. This process adaptively shifts the search target from temporal regions to specific objects, and visual details to help video understanding. Experiments on VideoMME, LongVideoBench, and LVBench show that $T^3$ improves Qwen2.5-VL-7B by 0.5%, 4.6%, and 4.4%, respectively, under the same frame budget, demonstrating the effectiveness of structured temporal reasoning. Our code is available at https://github.com/hufflepuff0596/Temporal-Tree-of-Thought.

## 1 Introduction

Multimodal Large Language Models (MLLMs) (Cheng et al., 2024; Zhang et al., 2024b; Bai et al., 2025; Chen et al., 2024; Shen et al., 2024; Nie et al., 2024) have achieved strong performance in video-language understanding. However, their effectiveness on long videos remains constrained by limited context length. For example, a 30-minute video can easily produce over 400K visual tokens, far exceeding the context budget of most current video MLLMs and making exhaustive frame encoding computationally prohibitive. This challenge has motivated agent-based video understanding methods (Yang et al., 2025; Wang et al., 2025b; Liu et al., 2025; Wang et al., 2024), which aim to identify sparse yet relevant visual evidence from the entire video.

Yet, videos are more than sequences of isolated frames. When answering a question about a long video, humans rarely inspect frames independently or uniformly. Instead, they typically follow a progressive search process: first locating the relevant temporal segment, then focusing on the objects, or visual details within that segment, and finally integrating the gathered evidence to answer the question, as in Figure 1. This process suggests two important properties of long-video understanding. First, videos exhibit temporal structure: frames are organized into coherent temporal segments rather than independent observations. Second, the search target is not fixed throughout reasoning. At a coarse stage, the goal is to identify which part of the video is relevant; at a finer stage, the goal shifts toward finding specific objects, or visual evidence needed to support the answer.

Motivated by these perspectives, we introduce Temporal Tree of Thought ($T^3$), a training-free framework for long-video understanding.

Offline Temporal Tree Construction. We first build a question-agnostic Temporal Tree of the video. The tree is built through recursive temporally constrained clustering, where each node corresponds to a contiguous temporal segment. Each segment is represented by the frame closest to its cluster center and is recursively subdivided into shorter contiguous subsegments until reaching individual frames. The resulting tree provides a multi-granularity organization of the video content, serving as a structured search space for subsequent

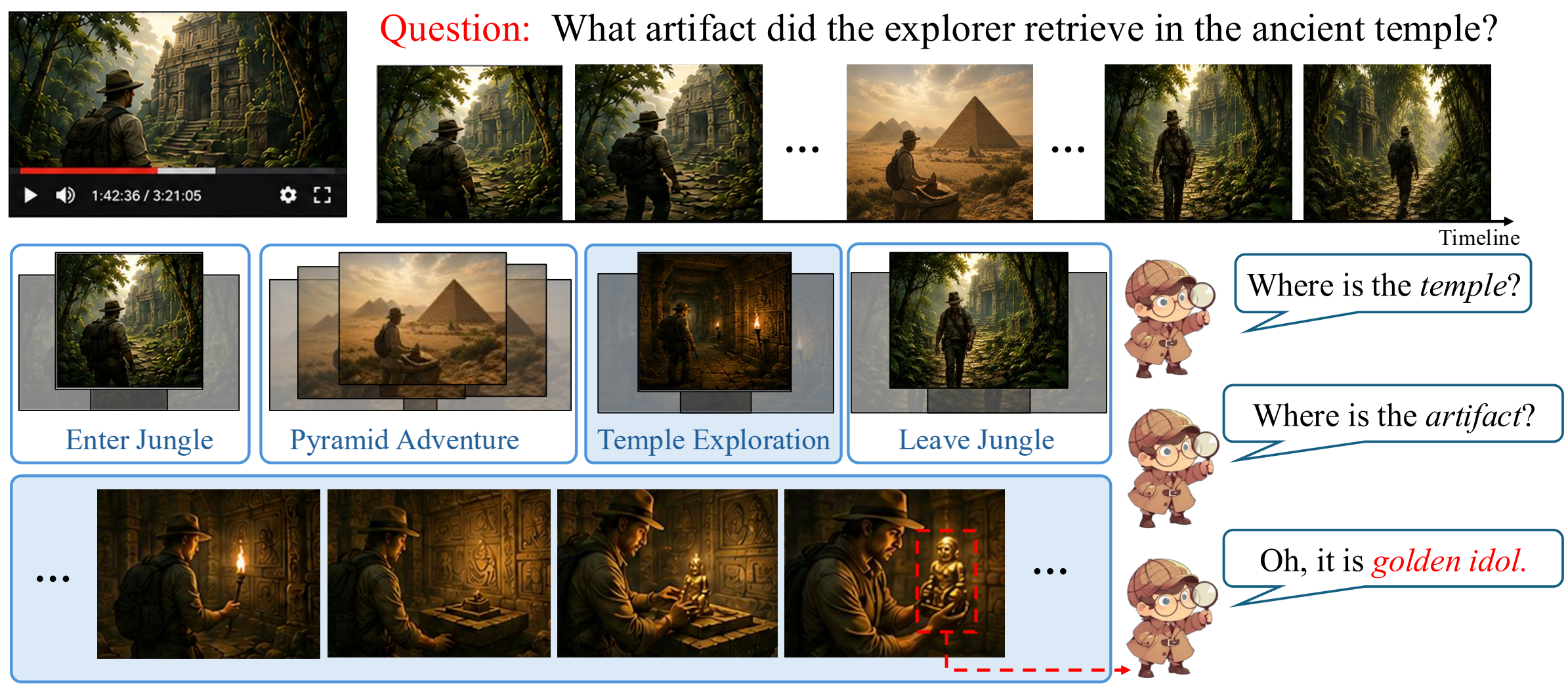

**Figure 1:** Illustration of progressive evidence search in long-video understanding. Rather than inspecting frames independently or uniformly, humans typically first locate the relevant temporal segment, e.g., the temple exploration segment, then zoom into local evidence such as objects, actions, and visual details, e.g., the retrieved artifact, before integrating the evidence to answer the question.

reasoning.

Tree-based Coarse-to-Fine Reasoning. At inference time, $T^3$ performs a progressive search over the constructed video tree. It starts from a coarse level, where representative frames provide a compact overview for locating potentially relevant temporal segments. Given the question and selected frames, the MLLM produces a tentative answer and a confidence score; if the confidence exceeds a calibrated threshold, reasoning terminates. Otherwise, the model generates a short search statement based on the current evidence, specifying what should be examined next. Since the available evidence becomes more detailed after each expansion, the search statement naturally evolves from coarse temporal cues to finer visual cues, such as objects, or local details. This answer–retrieve–explore loop continues until sufficient evidence is collected or the maximum number of rounds is reached.

We evaluate our method on three long-video QA benchmarks: VideoMME (Fu et al., 2025a), LongVideoBench (Wu et al., 2024), and LVBench (Wang et al., 2025a). Using Qwen2.5-VL-7B as the base model, our approach brings accuracy gains of 0.5%, 4.6%, and 4.4% on these benchmarks, respectively. These results show that tree-guided online reasoning can consistently improve video understanding, with particularly noticeable benefits on harder long-video benchmarks.

In summary, our contributions are as follows:

- We introduce a hierarchical temporal representation for long videos. Our method organizes videos into coarse-to-fine contiguous temporal segments through recursive clustering, providing a compact yet expandable representation of long video content.
- We propose $T^3$, a training-free framework for structured long-video reasoning. $T^3$ performs inference through an answer–retrieve–explore loop, adaptively selecting question-relevant branches of the temporal tree without additional training or fine-tuning.
- Our method achieves improvements of 0.5%, 4.6%, and 4.4% over the base model Qwen2.5-VL-7B on VideoMME, LongVideoBench, and LVBench, respectively.

## 2 Related Works

### 2.1 Long Video Understanding with MLLMs

Recent advances in Video Multimodal Large Language Models (Video MLLMs) have unified diverse video understanding tasks such as captioning, question answering, and retrieval (Cheng et al., 2024; Zhang et al., 2024b; Chen et al., 2024; Bai et al., 2025; Shen et al., 2024; Li et al., 2024a; Zhang et al., 2024a; Li et al., 2024b; Shu et al., 2025; Ren et al., 2025; Li et al., 2025; Nie et al., 2024). The Agent-based methods treat long-video understanding as an interactive reasoning process (Wang et al., 2024; Liu et al., 2025; Shen et al., 2025; Wang et al., 2025b). VideoAgent (Wang et al., 2024) performs multi-round decision making with an MLLM agent that retrieves captions or objects and verifies answers iteratively. VideoMind (Liu et al., 2025) extends this idea with role-based agents (planner, grounder, verifier, answerer) for long video reasoning.

Although VideoTree (Wang et al., 2025b) and other tree-based method (Cao et al., 2025) methods also use a hierarchical representation for long-video reasoning, $T^3$ introduces adaptivity differently. VideoTree builds a query-adaptive tree for each question by clustering visual features and evaluating candidate clusters against the query. In contrast, $T^3$ builds a reusable question-agnostic temporal tree with temporal constraints, so each node remains a contiguous video segment that preserves chronological structure.

Compared with other agent-based methods (Yang et al., 2025; Wang et al., 2025b; Nie et al., 2024; Fu et al., 2025b), $T^3$ performs goal-directed search over a structured temporal tree instead of repeatedly querying the entire video with the same original question. At each reasoning step, the MLLM updates the search statement according to the current evidence and tentative answer, enabling the retrieval target to shift from coarse temporal regions to fine-grained objects, and visual details. This progressive query refinement mitigates the risk of being trapped in a narrow evidence region and supports more effective coarse-to-fine evidence exploration.

### 2.2 Multimodal Reasoning

Multimodal reasoning with large models has evolved from single-pass prompting to deliberate, multi-step procedures. Chain-of-Thought (CoT) prompting elicits step-by-step rationales that improve interpretability and reasoning depth (Wei et al., 2022), while self-consistency enhances robustness by aggregating multiple reasoning paths (Wang et al., 2023). Tree-of-Thought (ToT) generalizes these ideas to a structured search over intermediate states, allowing look-ahead and backtracking instead of committing to a single reasoning chain (Yao et al., 2023). In parallel, reasoning-and-acting frameworks interleave planning with tool usage, such as retrieval, computation, or vision modules with ReAct serving as a canonical paradigm that alternates between thoughts, actions, and observations (Yao et al., 2022).

## 3 Temporal Tree of Thought

We study training-free multiple-choice question answering on long videos with Multimodal Large Language Models. Let a video be a sequence of frames $v = \{f_1, \ldots, f_N\}$ and a question $q$ with candidate options $\mathcal{O} = \{o_1, \ldots, o_M\}$. The goal is to predict the correct option $o_\star \in \mathcal{O}$. An overview of our framework is shown in Figure 2.

### 3.1 Temporal Tree

Before inference, we construct a query-agnostic temporal tree that organizes the video along the temporal dimension, serving as a compact external memory and access path to the raw frames. Given a video $v = \{f_1, \ldots, f_N\}$, we first densely sample frames and extract visual features using a frozen CLIP image encoder (Radford et al., 2021):

$$x_i = \phi(f_i) \in \mathbb{R}^d, \quad i = 1, \ldots, N,$$

where $\phi(\cdot)$ denotes the image encoder. These features are computed once and cached for later use in indexing and retrieval, keeping the entire indexing process training-free.

We then organize the video into a hierarchical temporal tree by recursively partitioning the timeline into contiguous segments with coherent visual features. At the top level ($\ell = 1$), the full video is divided into $K_1$ coarse segments via temporal clustering in the feature space, producing segments that summarize broad temporal regions. Each segment is represented by a representative frame: the frame whose feature is closest to the cluster center which serves as the entry for that branch. At the next level ($\ell = 2$), each segment from level $\ell = 1$ is further subdivided into $K_2$ finer, contiguous subsegments using the same rule, again assigning a representative frame to each child. This recursive partitioning continues until a segment contains fewer than $L_{\min}$ frames (we set $L_{\min} = 1$), yielding a multi-level temporal tree. Formally, the resulting hierarchical tree is denoted as

$$\mathcal{T} = \{R_{(1)}, \ldots, R_{(L)}\}, \quad R_{(\ell)} = \{r_1, \ldots, r_{N_\ell}\},$$

where $R_{(\ell)}$ denotes the set of nodes at level $\ell$. Each $r_{\ell j}$ stores the representative frame, its visual feature, the corresponding time span, and pointers to its children, so higher levels provide coarse, lightweight summaries while deeper levels offer progressively finer access. The clustering code is provided in Appendix.

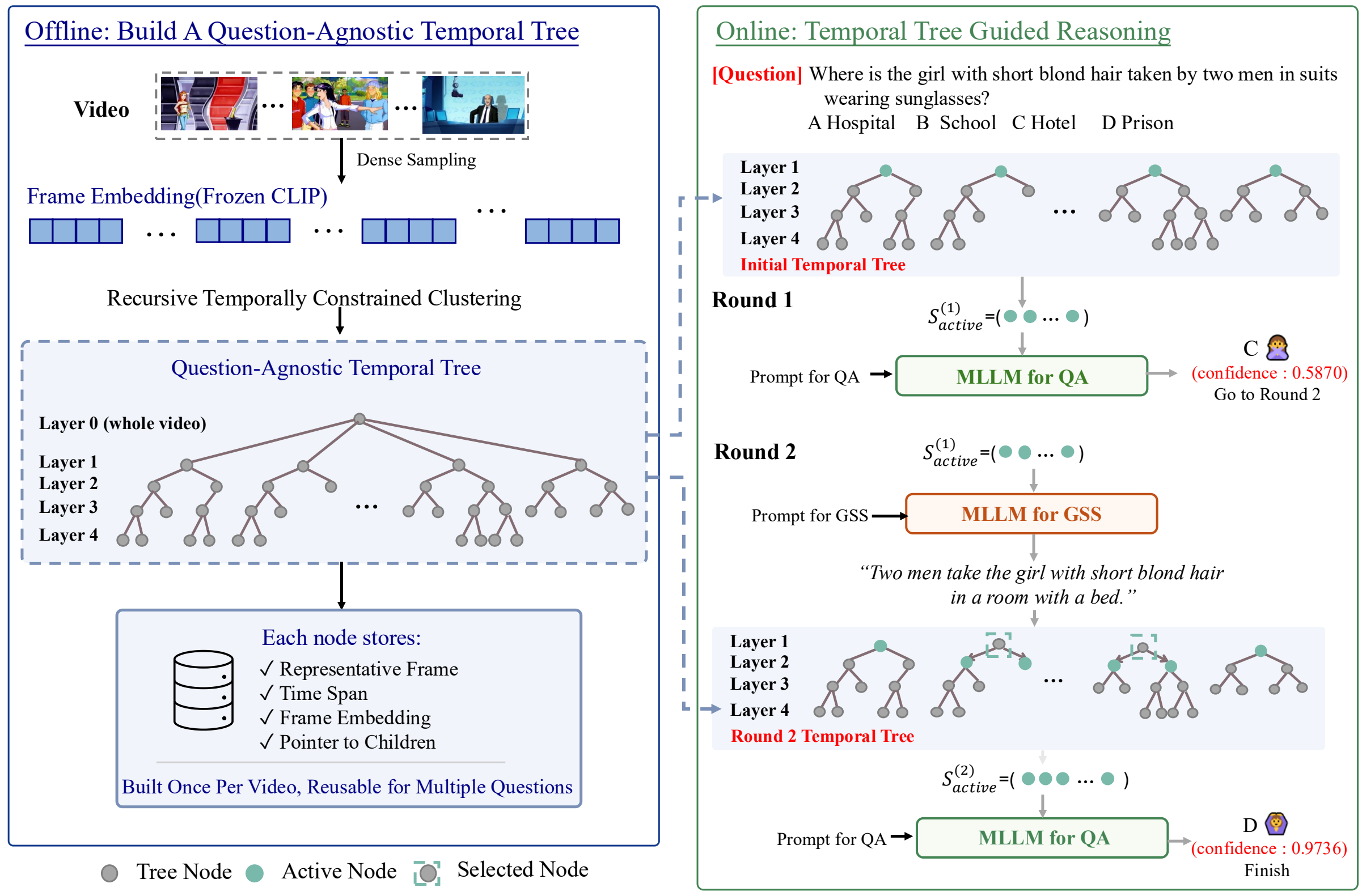

**Figure 2:** Framework of $T^3$ We first build a query-agnostic temporal tree, where each node contains multiple information,but we only use representative frame for reasoning. In Reasoning Round 1, a frozen MLLM predicts an answer from coarse video information. If its confidence is low, $T^3$ enters Round 2: it generates a Generate Search Sentence (GSS) to describe what should be examined next. Only two rounds are shown; Round 2 can repeat for further refinement.

### 3.2 Temporal Tree Guided Reasoning

At inference time, given a question $q$, we run a multi-round retrieve–reason–explore loop based on temporal tree.

**Initial reasoning.** Instead of directly scoring all frames, the model always operates over a small set of representative frames, starting from first layer representatives that cover the whole timeline and gradually refining only when the current evidence is insufficient. These representative frames collectively provide global coverage over the entire timeline and form the initial active set $S^1_{\text{active}} = R_{(1)}$. We feed $S^1_{\text{active}}$ together with the question $q$ and the answer options $\mathcal{O}$ into the frozen MLLM to obtain an initial prediction. The MLLM outputs a predicted option $o^1_\star \in \mathcal{O}$ and a confidence score $c^1$, defined as the next-token probability assigned to the corresponding option letter:

$$c^1 = p(o^1 \mid S^1_{\text{active}}, q).$$

If $c^1 \geq \tau$, the model is already confident given the coarse global context and we terminate with $o^1$ as the final answer. Otherwise, a low confidence score indicates that the current representative frames do not provide sufficient evidence at this granularity, and we proceed to subsequent rounds to refine the active set by seeking additional visual evidence.

**Generate Search Sentence and Guided Retrieval.** When $c^1 <= \tau$, the current evidence is insufficient for a confident decision. We therefore prompt the frozen MLLM to generate a concise search sentence $u^2$ to state what should be examined next. This sentence plays the role of a retrieval query that find next relevant objects. We embed $u^2$ using a frozen CLIP text encoder $\psi(\cdot)$ and score it against each active representative frames $r \in S^1_{\text{active}}$ via cosine similarity:

$$s(u^2, r) = \frac{\langle \psi(u^2), \phi(r) \rangle}{\|\psi(u^2)\|_2 \, \|\phi(r)\|_2}, \qquad r \in S^1_{\text{active}}.$$

Let $s_{\max} = \max_{r \in S^1_{\text{active}}} s(u^2, r)$. We select representative frames whose similarity satisfies $s(u^2, r) \geq \rho \, s_{\max}$, forming the retrieved subset $S^2_{\text{selected}}$. We use CLIP similarity only as a lightweight retrieval signal to localize candidate representative frames. Once a representative frames are selected and expanded, the MLLM receives multiple representative frames from finer level, enabling it to reason over more details.

**Selective Exploration of Temporal Tree Branch.** We then refine only the retrieved regions by replacing with the representative frames in the next layer. Concretely, for each selected representative frames $r \in S^2_{\text{selected}}$, we fetch its children $\text{Children}(r)$ from the next level and update the active set by replacing the selected parents with their children:

$$S^2_{\text{active}} = \left(S^1_{\text{active}} \setminus S^2_{\text{selected}}\right) \cup \text{Children}\left(S^2_{\text{selected}}\right).$$

If the queried cue (e.g., "red ball") is absent from all current representative frames, similarities are uniformly low and close to each other, and the ratio rule selects a broader set of entries for expansion. In this case, the method naturally widens its search over multiple finer branches and continues refining until the evidence emerges at a finer level or the maximum number of rounds is reached.

**Multi-Round Reasoning.** The above answer–retrieve–explore steps are executed iteratively for $t = 1, \ldots, T_{\max}$. At each round, the frozen MLLM takes the current active set $S^t_{\text{active}}$ together with the question $q$ to produce a prediction $o^t_\star$ and its confidence $c^t$. The process terminates early if $c^t \geq \tau$, since the current retrieved evidence is deemed sufficient, or stops once the maximum number of rounds $T_{\max}$ is reached. Otherwise, the model generates a new evidence query $u^{t+1}$, uses it to re-score the active representative frames and select a subset $S^{t+1}_{\text{selected}}$ for exploration, and updates the active set via selective expansion as above. In the end, we output the prediction from the final round as the answer. Because $S^t_{\text{active}}$ always contains representative frames distributed across the timeline, each round maintains global coverage: the model may either drill down within a promising region or select other related representative frames to explore new story regions.

# 4 Experiments

**Implementation details.** We evaluate $T^3$ with two representative multimodal LLM backbones: Qwen2.5-VL-7B (Bai et al., 2025) and LLaVA-Video-7B (Zhang et al., 2024b); all ablation studies are conducted on Qwen2.5-VL-7B. In the offline stage, videos are densely sampled at 2 fps, and frame-level visual embeddings are extracted using a frozen CLIP ViT-B/32 (Radford et al., 2021) encoder. All features are cached to avoid repeated video decoding. A query-agnostic temporal tree is then constructed by temporal clustering: the top level contains $K_1$=64 coarse segments, and each segment is recursively subdivided into $K_{>1}$=4 child segments. In the online stage, we perform multi-round reasoning for at most $T_{\max}$=3 rounds on VideoMME (Fu et al., 2025a) and LongVideoBench (Wu et al., 2024), and $T_{\max}$=5 rounds on LVBench (Wang et al., 2025a) because different average temporal tree depth. At each round, inference terminates early if the next-token confidence $c \geq 0.9$; otherwise, we expand only nodes whose CLIP text–image similarity satisfies $s \geq \rho\, s_{\max}$, with $\rho$=0.75 by default. The frames processed by all rounds are within 128 frames.

**Datasets.** We evaluate our method on three long-video multiple-choice QA benchmarks. VideoMME (Fu et al., 2025a) is a human-annotated evaluation dataset covering videos from roughly 11 s to 1 h in length, spanning six domains and 30 subcategories. LongVideoBench (Wu et al., 2024) is a large-scale long-context QA benchmark built from web videos with subtitles, featuring interleaved video–text inputs up to one hour per sample. It contains 3,763 videos and 6,678 human-written questions across diverse real-world themes; we report results on the official validation set without subtitle. LVBench (Wang et al., 2025a) focuses on extremely long videos collected from public YouTube sources such as sports, live streams, documentaries, and animations. Its average video duration is approximately 4,101 s (∼68.4 min), substantially longer than those in prior benchmarks, therefore offering a more challenging evaluation setting for long video understanding.

## 4.1 Main Results

In Table 1, we compare the proposed $T^3$ with recent agent-based methods (Wang et al., 2025b; Yang et al., 2025; Luo et al., 2024), as well as their backbone models, including Qwen2.5-VL-7B and LLaVA-Video-7B. For prior methods, we follow their original reported settings, including the frame budget, to ensure that each method is evaluated under its intended configuration.

**VideoMME (w/o sub).** $T^3$-Q reaches 65.4 overall, improving upon Qwen2.5-VL (64.9) by +0.5, while $T^3$-L improves LLaVA-Video from 61.4 to 63.7 (+2.3). On the long-video subset (30–60 min), the gains remain consistent: $T^3$-Q increases from 53.9 to 55.3 (+1.4) and $T^3$-L from 52.9 to 54.2 (+1.3), suggesting that structured guided reasoning becomes more beneficial as video length grows. Compared with recently reported multi-round reasoning baselines on VideoMME (e.g., VideoTree), $T^3$ achieves competitive performance. Moreover, under the same Qwen2.5-VL-7B backbone and same frame budget, $T^3$-Q outperforms Video-RAG (62.1 overall), highlighting the effectiveness of structured navigation for evidence localization.

| Models | Size | #Frames | VideoMME (w/o sub) | | LongVideoBench | LVBench |
|---|---|---|---|---|---|---|
| | | | Long | Overall | | |
| *Duration* | | | 30–60 min | 0–60 min | 0–60 min | 4,101 s |
| ***Proprietary Video MLLMs*** | | | | | | |
| GPT4o (Hurst et al., 2024) | – | 1fps | 77.2 | 72.1 | 66.7 | 34.7 |
| Gemini-1.5-Pro (Google, 2024) | – | – | – | 75.7 | 62.9 | – |
| VCA (Yang et al., 2025) | – | – | – | – | 41.3 | – |
| VideoTree (Wang et al., 2025b) | – | – | 54.2 | – | – | – |
| ***Open-Source Video MLLMs*** | | | | | | |
| Video-LLaVA (Lin et al., 2024) | 7B | 8 | 36.2 | 39.9 | 39.1 | – |
| ShareGPT4Video (Chen et al., 2024) | 8B | 16 | 33.2 | 39.5 | 39.7 | – |
| LongVA (Zhang et al., 2024a) | 7B | 128 | 47.6 | 54.3 | – | – |
| Vamba (Ren et al., 2025) | 10B | 1024 | – | 57.8 | 55.9 | 42.1 |
| VideoMind (Liu et al., 2025) | 7B | – | 58.2 | 49.2 | 56.3 | 40.8 |
| Video-R1 (Feng et al., 2025) | 7B | 64 | – | 61.4 | – | – |
| LongVILA-R1 (Chen et al., 2025) | 7B | 512 | – | 65.1 | 58.0 | – |
| Video-RAG (Luo et al., 2024) | 7B | 128 | 59.8 | 62.1 | 58.7 | – |
| Qwen2.5-VL† (Bai et al., 2025) | 7B | 768 | – | 65.1 | 54.7 | 45.3 |
| LLaVA-Video* (Zhang et al., 2024b) | 7B | 128 | 52.9 | 61.4 | 55.7 | 45.2 |
| $T^3$-L (Ours) | 7B | 128 | 54.2 +1.3 | 63.7 +2.3 | 61.4 +5.7 | 47.3 +2.1 |
| Qwen2.5-VL* (Bai et al., 2025) | 7B | 128 | 53.9 | 64.9 | 56.0 | 42.0 |
| $T^3$-Q (Ours) | 7B | 128 | 55.3 +1.4 | 65.4 +0.5 | 60.6 +4.6 | 46.4 +4.4 |

**Table 1:** Evaluation results on long-video understanding benchmarks. #Frames denotes frames budget per sample. *: reproduced by us. †: Official reported. $T^3$-L is based on LLaVA-Video; $T^3$-Q is based on Qwen2.5-VL. VideoTree and VCA are based on GPT-4o (Hurst et al., 2024).

**LongVideoBench.** Larger gains are observed on LongVideoBench, which emphasizes temporally localized evidence. $T^3$-Q improves from 56.0 to 60.6 (+4.6), and $T^3$-L from 55.7 to 61.4 (+5.7), indicating that multi-round reasoning over a hierarchical temporal tree is effective when relevant evidence is sparse and dispersed across the timeline. Compared with multi-round reasoning baselines such as VCA and retrieval-augmented methods such as Video-RAG (58.7), $T^3$ achieves strong performance. Notably, $T^3$ also remains competitive despite using far fewer frames than densely sampled models (e.g., Vamba with 1024 frames), highlighting the efficiency of our proposed method.

**LVBench.** On LVBench (average length 4,101 seconds), $T^3$-Q improves from 42.0 to 46.4 (+4.4), and $T^3$-L from 45.2 to 47.3 (+2.1). These gains demonstrate the effectiveness of proposed $T^3$.

We attribute the smaller overall gain on VideoMME to its duration distribution and question composition. First, VideoMME contains short, medium, and long videos, with an average duration of about 17 minutes. For shorter videos, uniformly sampling 128 frames already provides relatively dense coverage, leaving less room for additional temporal search. Consistently, on the 30–60 minute subset, $T^3$ improves Qwen2.5-VL from 53.9 to 55.3 (+1.4), compared with the +0.5 overall gain. Second, many VideoMME questions require global aggregation or coarse video understanding rather than sparse evidence localization. Such ordering, counting, and summary questions are already reasonably handled by broad uniform sampling. In contrast, LongVideoBench emphasizes referred reasoning over specific temporal contexts, which better matches $T^3$'s coarse-to-fine search and explains the larger +4.6 gain.

## 4.2 Ablation Studies

**Model Design.** Table 2 reports an ablation study on VideoMME to quantify the contribution of each component in our video indexing and reasoning framework. We start from a naive Uniform Sampling baseline that uniformly samples 64 frames, ignoring temporal structure.

| Variant | Long | Overall | Δ |
|---|---|---|---|
| Uniform Sampling | 51.56 | 61.93 | — |
| + Structure Aware Representation | 51.56 | 63.04 | +1.11 |
| + Question Guided Retrieval | 52.78 | 63.67 | +1.74 |
| + Reasoning Guided Retrieval | 53.11 | 64.15 | +2.22 |
| + Adaptive Expansion | **55.33** | **65.44** | **+3.51** |
| Non-Structured Reasoning | 52.00 | 63.19 | — |
| Non-Temporal Constraint | 51.67 | 63.33 | — |

**Table 2:** Ablation study on VideoMME evaluating the impact of temporal tree and multi-round reasoning. Each row incrementally adds a design component to the uniform sampling baseline, isolating the effects of structure-aware representation, retrieval guidance, and adaptive expansion.

Replacing sampled frames with a Structure-Aware Representation where using the 64 first level representative frames, already yields a clear improvement, around +1.11 overall, indicating that a coarse temporal overview provides more informative context than flat frame selection. Building on this structure, Question-Guided Retrieval uses the question text as a fixed CLIP query to iteratively expand the Top-2 most relevant segments, further improving performance (+0.63), suggesting that coarse alignment between the fixed query helps find useful cues to answer the question. We then introduce Reasoning-Guided Retrieval, where the model generates a round-specific search sentence that explicitly describes what to explore next; this produces more precise retrieval signals and adds another +0.48 gain, which means updated retrieval query can further improve focus. Finally, Adaptive Expansion replaces hard Top-$k$ expansion with a ratio-based rule that expands all segments whose similarity exceeds $\rho\, s_{\max}$ ($\rho$=0.75), achieving the largest boost (+1.29 over reasoning-guided retrieval and +3.51 over uniform sampling). This adaptive multi-branch refinement is particularly important for long videos, where relevant evidence may be distributed across multiple temporal regions; when the cue is missing from current active set and similarities are uniformly low, the ratio rule expands more branches to widen the search and continue exploring.

As a contrast, Non-Structured Reasoning, which builds a binary temporal tree via uniform partitioning without leveraging temporally coherent segmentation, which is used in most of agent-based methods (Yang et al., 2025; Shen et al., 2025), performs substantially worse than the structured variant, highlighting the importance of an explicit temporal tree for effective hierarchical reasoning on long videos. We further evaluate a Non-Temporal Constraint variant, which removes the temporal constraint during clustering, which is used in VideoTree (Wang et al., 2025b), regardless of their chronological order. This variant underperforms the full structure-aware setting, indicating that preserving temporal contiguity is important: visually similar but temporally distant frames may correspond to different events or contexts, and grouping them together can weaken the tree as a meaningful search space.

**Effect of Stepwise Reasoning.** Table 3 summarizes how multi-round reasoning affects evidence recovery on LVBench. Among 1,509 questions with annotated reference spans, 12.2% terminate after a single round, while 87.8% proceed for multiple rounds. Nearly half of the multi-round cases (46.3%) exhibit improved target coverage, indicating that refinement frequently expands the retrieved region toward the ground-truth evidence. Within this subset, 22.0% are late-hit cases where Round 1 covers no target frames but later rounds (Rounds 2–5) successfully retrieve them. Accuracy improves in parallel: 250 samples are already correct after Round 1, increasing to 314 after Round 5, a net gain of 64 correct answers (+4.2 points overall; +25.6% relative to Round 1).

| Category | Number | % |
|---|---|---|
| **Single-Round** | 184 | 12.2 |
| **Multi-Round** | 1325 | 87.8 |
| → Improved | 699 | 46.3 |
| Late-Hit | 332 | 22.0 |
| Correct @ Round 1 | 250 | 16.6 |
| Correct @ Round 5 | 314 | 20.8 |
| → w/o Improved | 626 | 41.5 |

**Table 3:** Breakdown of step reasoning on LVBench. "Improved" indicates cases where later refinement (Round 2–5) captures more frames in ground truth than the first step (Round 1); "Late-Hit" refers to samples where the initial round misses all targets but subsequent reasoning recovers them.

The remaining 41.5% of multi-round cases without coverage gains likely correspond to questions where the initial retrieval already captures the key moment or where the text–image matching signal is insufficient to introduce additional evidence. Overall, these results support the utility of multi-round reasoning: when initial context is insufficient, additional rounds often recover missing evidence and translate it into measurable accuracy gains.

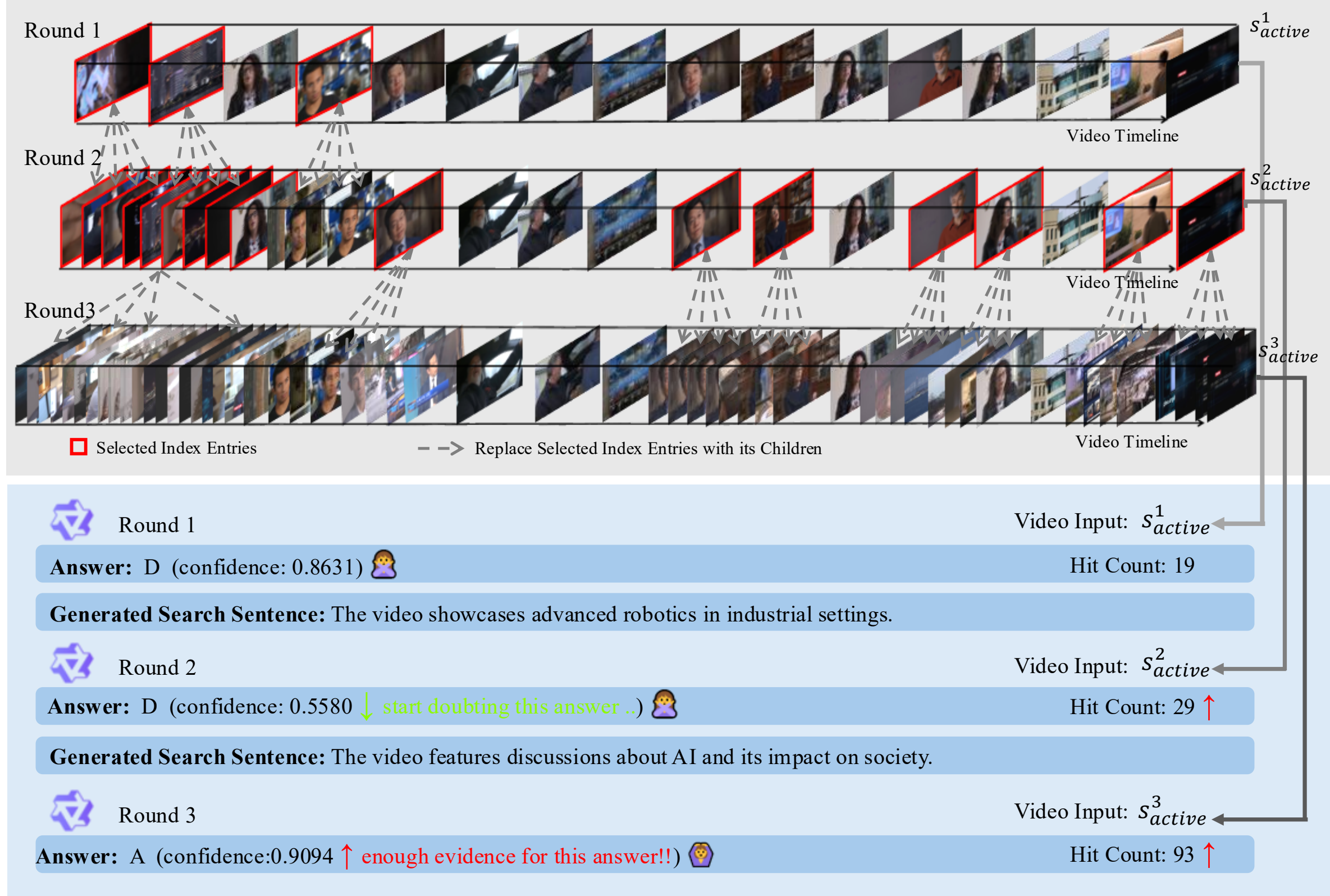

**Figure 3:** Visualization of step reasoning on LVBench. Example from the $T^3$-Q model showing how frame selection refines across rounds 1–3. Hit Count denotes the number of selected frames overlapping the annotated evidence indices. For clearer illustration, we uniformly sample 16 representative frames from the 64 first-layer representative frames.

**Qualitative Visualization.** Figure 3 presents a qualitative LVBench example showing how $T^3$-Q revises both where it looks and what it looks for (GSS). Crucially, $T^3$ does not commit to a single branch after Round 1: the active set always retains representative frames distributed across the full timeline, so the model can drill down into a candidate region or shift to other stories when evidence is inconsistent. In Round 1, $T^3$ scans coarse representative frames and predicts 'D' with high confidence (0.8631). Based on the current context, it hypothesizes the video is about 'advanced robotics', and the generate search sentence reflects this cue. In Round 2, $T^3$ expands and re-scores representative frames guided by the robotics cue, which increases overlap with the reference span (hit count 29). Yet the confidence for 'D' drops to 0.5580, indicating that the additional frames introduce competing cues and weaken the robotics interpretation. $T^3$ therefore updates the retrieval key: the new search sentence shifts to 'AI and its societal impact', explicitly reflecting what it now believes is the missing evidence. In Round 3, guided by the updated AI-focused query, $T^3$ navigates to alternative temporal regions and retrieves substantially denser supporting evidence (hit count 93). With sufficient support collected, the prediction switches to 'A' and confidence rises to 0.9094, at which point $T^3$ terminates. Overall, this example highlights $T^3$'s ability to move beyond an initially plausible but weak hypothesis, update the search cue, and recover missing evidence through multi-round reasoning.

## 5 Conclusion

We presented $T^3$, a training-free framework for long-video understanding that organizes each video into a query-agnostic temporal tree and performs multi-round online reasoning over it. Starting from coarse representative frames, the model identifies missing visual evidence and selectively explores relevant temporal regions. This decouples temporal tree construction from question-specific reasoning, allowing reuse across questions. Experiments on VideoMME, LongVideoBench, and LVBench show that $T^3$ consistently improves long-video question answering. Ablations validate the contributions of temporal representation, and adaptive expansion. Overall, combining a reusable temporal tree with multi-round reasoning offers an effective direction for long-video understanding with MLLMs.

## 6 Limitations

Despite its effectiveness, $T^3$ has several limitations that suggest promising directions for future research.

First, the current temporal tree is constructed purely from visual features, without incorporating other available modalities such as subtitles, audio, or ASR transcripts. While this design keeps the temporal tree construction stage simple and training-free, integrating multimodal cues could lead to segment boundaries that better align with semantic transitions, especially in dialogue-heavy or audio-centric videos.Second, each temporal segment is represented by a single visual representative frame. Although sufficient for coarse searching and object-level retrieval, this representation may miss whole some cues like motion within a segment. A natural extension is to replace single-frame with richer segment-level summaries, such as compact textual descriptions that describe what happens within each segment.Moreover, Our current retrieval module uses CLIP image-text similarity between the generated search statement and representative frames. This design is lightweight and efficient, but it may under-represent motion patterns and temporal relations that cannot be captured by a single frame.

We believe these directions will further strengthen the role of structured indexing and evidence-driven inference in long-video understanding.

## References

Shuai Bai, Keqin Chen, Xuejing Liu, Jialin Wang, Wenbin Ge, Sibo Song, Kai Dang, Peng Wang, Shijie Wang, Jun Tang, Humen Zhong, Yuanzhi Zhu, Mingkun Yang, Zhaohai Li, Jianqiang Wan, Pengfei Wang, Wei Ding, Zheren Fu, Yiheng Xu, and 8 others. 2025. Qwen2. 5-vl technical report. *arXiv:2502.13923*.

Xinye Cao, Hongcan Guo, Jiawen Qian, Guoshun Nan, Chao Wang, Yuqi Pan, Tianhao Hou, Xiaojuan Wang, and Yutong Gao. 2025. Videominer: Iteratively grounding key frames of hour-long videos via tree-based group relative policy optimization. In *ICCV*.

Lin Chen, Xilin Wei, Jinsong Li, Xiaoyi Dong, Pan Zhang, Yuhang Zang, Zehui Chen, Haodong Duan, Lin Bin, Zhenyu Tang, Li Yuan, Qiao Yu, Dahua Lin, Feng Zhao, and Jiaqi Wang. 2024. Sharegpt4video: Improving video understanding and generation with better captions. In *NeurIPS*.

Yukang Chen, Wei Huang, Baifeng Shi, Qinghao Hu, Hanrong Ye, Ligeng Zhu, Zhijian Liu, Pavlo Molchanov, Jan Kautz, Xiaojuan Qi, Sifei Liu, Hongxu Yin, Yao Lu, and Song Han. 2025. Scaling rl to long videos. In *NeurIPS*.

Zesen Cheng, Sicong Leng, Hang Zhang, Yifei Xin, Xin Li, Guanzheng Chen, Yongxin Zhu, Wenqi Zhang, Ziyang Luo, Deli Zhao, and Lidong Bing. 2024. Videollama 2: Advancing spatial-temporal modeling and audio understanding in video-llms. *arXiv:2406.07476*.

Kaituo Feng, Kaixiong Gong, Bohao Li, Zonghao Guo, Yibing Wang, Tianshuo Peng, Junfei Wu, Xiaoying Zhang, Benyou Wang, and Xiangyu Yue. 2025. Video-r1: Reinforcing video reasoning in mllms. *arXiv:2503.21776*.

Chaoyou Fu, Yuhan Dai, Yongdong Luo, Lei Li, Shuhuai Ren, Renrui Zhang, Zihan Wang, Chenyu Zhou, Yunhang Shen, Mengdan Zhang, and 1 others. 2025a. Video-mme: The first-ever comprehensive evaluation benchmark of multi-modal llms in video analysis. In *CVPR*.

Shenghao Fu, Qize Yang, Yuan-Ming Li, Xihan Wei, Xiaohua Xie, and Wei-Shi Zheng. 2025b. Love-r1: Advancing long video understanding with an adaptive zoom-in mechanism via multi-step reasoning. *arXiv:2509.24786*.

Gemini Team Google. 2024. Gemini 1.5: Unlocking multimodal understanding across millions of tokens of context. *arXiv:2403.05530*.

Aaron Hurst, Adam Lerer, Adam P Goucher, Adam Perelman, Aditya Ramesh, Aidan Clark, AJ Ostrow, Akila Welihinda, Alan Hayes, Alec Radford, and 1 others. 2024. Gpt-4o system card. *arXiv:2410.21276*.

Bo Li, Yuanhan Zhang, Dong Guo, Renrui Zhang, Feng Li, Hao Zhang, Kaichen Zhang, Yanwei Li, Ziwei Liu, and Chunyuan Li. 2024a. Llava-onevision: Easy visual task transfer. *arXiv:2408.03326*.

KunChang Li, Yinan He, Yi Wang, Yizhuo Li, Wenhai Wang, Ping Luo, Yali Wang, Limin Wang, and Yu Qiao. 2025. Videochat: Chat-centric video understanding. In *Science China Information Sciences*.

Yanwei Li, Chengyao Wang, and Jiaya Jia. 2024b. Llama-vid: An image is worth 2 tokens in large language models. In *ECCV*.

Bin Lin, Yang Ye, Bin Zhu, Jiaxi Cui, Munan Ning, Peng Jin, and Li Yuan. 2024. Video-llava: Learning united visual representation by alignment before projection. In *EMNLP*.

Ye Liu, Kevin Qinghong Lin, Chang Wen Chen, and Mike Zheng Shou. 2025. Videomind: A chain-of-lora agent for long video reasoning. *arXiv:2503.13444*.

Yongdong Luo, Xiawu Zheng, Guilin Li, Shukang Yin, Haojia Lin, Chaoyou Fu, Jinfa Huang, Jiayi Ji, Fei Chao, Jiebo Luo, and Rongrong Ji. 2024. Video-rag: Visually-aligned retrieval-augmented long video comprehension. *arXiv:2411.13093*.

Ming Nie, Dan Ding, Chunwei Wang, Yuanfan Guo, Jianhua Han, Hang Xu, and Li Zhang. 2024. Slowfocus: Enhancing fine-grained temporal understanding in video llm. In *NeurIPS*.

Alec Radford, Jong Wook Kim, Chris Hallacy, Aditya Ramesh, Gabriel Goh, Sandhini Agarwal, Girish Sastry, Amanda Askell, Pamela Mishkin, Jack Clark, Gretchen Krueger, and Ilya Sutskever. 2021. Learning transferable visual models from natural language supervision. In *ICML*.

Weiming Ren, Wentao Ma, Huan Yang, Cong Wei, Ge Zhang, and Wenhu Chen. 2025. Vamba: Understanding hour-long videos with hybrid mamba-transformers. In *ICCV*.

Xiaoqian Shen, Yunyang Xiong, Changsheng Zhao, Lemeng Wu, Jun Chen, Chenchen Zhu, Zechun Liu, Fanyi Xiao, Balakrishnan Varadarajan, Florian Bordes, Zhuang Liu, Hu Xu, Hyunwoo J. Kim, Bilge Soran, Raghuraman Krishnamoorthi, Mohamed Elhoseiny, and Vikas Chandra. 2024. Longvu: Spatiotemporal adaptive compression for long video-language understanding. *arXiv:2410.17434*.

Xiaoqian Shen, Wenxuan Zhang, Jun Chen, and Mohamed Elhoseiny. 2025. Vgent: Graph-based retrieval-reasoning-augmented generation for long video understanding. In *NeurIPS*.

Yan Shu, Zheng Liu, Peitian Zhang, Minghao Qin, Junjie Zhou, Zhengyang Liang, Tiejun Huang, and Bo Zhao. 2025. Video-xl: Extra-long vision language model for hour-scale video understanding. In *CVPR*.

Weihan Wang, Zehai He, Wenyi Hong, Yean Cheng, Xiaohan Zhang, Ji Qi, Ming Ding, Xiaotao Gu, Shiyu Huang, Bin Xu, Yuxiao Dong, and Jie Tang. 2025a. Lvbench: An extreme long video understanding benchmark. In *ICCV*.

Xiaohan Wang, Yuhui Zhang, Orr Zohar, and Serena Yeung-Levy. 2024. Videoagent: Long-form video understanding with large language model as agent. In *ECCV*.

Xuezhi Wang, Jason Wei, Dale Schuurmans, Quoc V Le, Ed Chi, Sharan Narang, Aakanksha Chowdhery, and Denny Zhou. 2023. Self-consistency improves chain-of-thought reasoning in language models. In *ICLR*.

Ziyang Wang, Shoubin Yu, Elias Stengel-Eskin, Jaehong Yoon, Feng Cheng, Gedas Bertasius, and Mohit Bansal. 2025b. Videotree: Adaptive tree-based video representation for llm reasoning on long videos. In *CVPR*.

Jason Wei, Xuezhi Wang, Dale Schuurmans, Maarten Bosma, Brian Ichter, Fei Xia, Ed Chi, Quoc V Le, and Denny Zhou. 2022. Chain-of-thought prompting elicits reasoning in large language models. In *NeurIPS*.

Haoning Wu, Dongxu Li, Bei Chen, and Junnan Li. 2024. Longvideobench: A benchmark for long-context interleaved video-language understanding. In *NeurIPS*.

Zeyuan Yang, Delin Chen, Xueyang Yu, Maohao Shen, and Chuang Gan. 2025. Vca: Video curious agent for long video understanding. In *ICCV*.

Shunyu Yao, Dian Yu, Jeffrey Zhao, Izhak Shafran, Tom Griffiths, Yuan Cao, and Karthik Narasimhan. 2023. Tree of thoughts: Deliberate problem solving with large language models. In *NeurIPS*.

Shunyu Yao, Jeffrey Zhao, Dian Yu, Nan Du, Izhak Shafran, Karthik R Narasimhan, and Yuan Cao. 2022. React: Synergizing reasoning and acting in language models. In *ICLR*.

Peiyuan Zhang, Kaichen Zhang, Bo Li, Guangtao Zeng, Jingkang Yang, Yuanhan Zhang, Ziyue Wang, Haoran Tan, Chunyuan Li, and Ziwei Liu. 2024a. Long context transfer from language to vision. *arXiv:2406.16852*.

Yuanhan Zhang, Jinming Wu, Wei Li, Bo Li, Zejun Ma, Ziwei Liu, and Chunyuan Li. 2024b. Video instruction tuning with synthetic data. *arXiv:2410.02713*.

## Appendix

# A Experiment

## A.1 Additional Experimental Setting

All evaluations are conducted on a single NVIDIA A100 GPU. For fair comparison, we follow the native resolution of each backbone, using an input image size of 448 for Qwen2.5-VL and 224 for LLaVA-Video. The maximum reasoning depth $T_{\text{max}}$ is set according to the average depth of the constructed video temporal tree on each dataset.

## A.2 Model Checkpoints

Table 4 lists the models we use and the checkpoints loaded during evaluation.

| Model name | Checkpoint |
|---|---|
| Qwen2.5–VL | `Qwen/Qwen2.5-VL-7B-Instruct` |
| LLaVA–Video | `lmms-lab/LLaVA-Video-7B-Qwen2` |
| CLIP | `ViT-B/32` |

**Table 4:** Model checkpoints used in our experiments.

## A.3 Examples of search sentences.

As shown in Table 5, the generated search sentence is not a simple rephrasing of the original question. Instead, it evolves according to the current retrieved evidence and intermediate prediction. Rather than simply zooming in, the generated search sentence explicitly specifies the missing evidence required to verify or revise the current hypothesis, which may either refine the current search or redirect it to a different temporal region.

## A.4 Task-Type Analysis

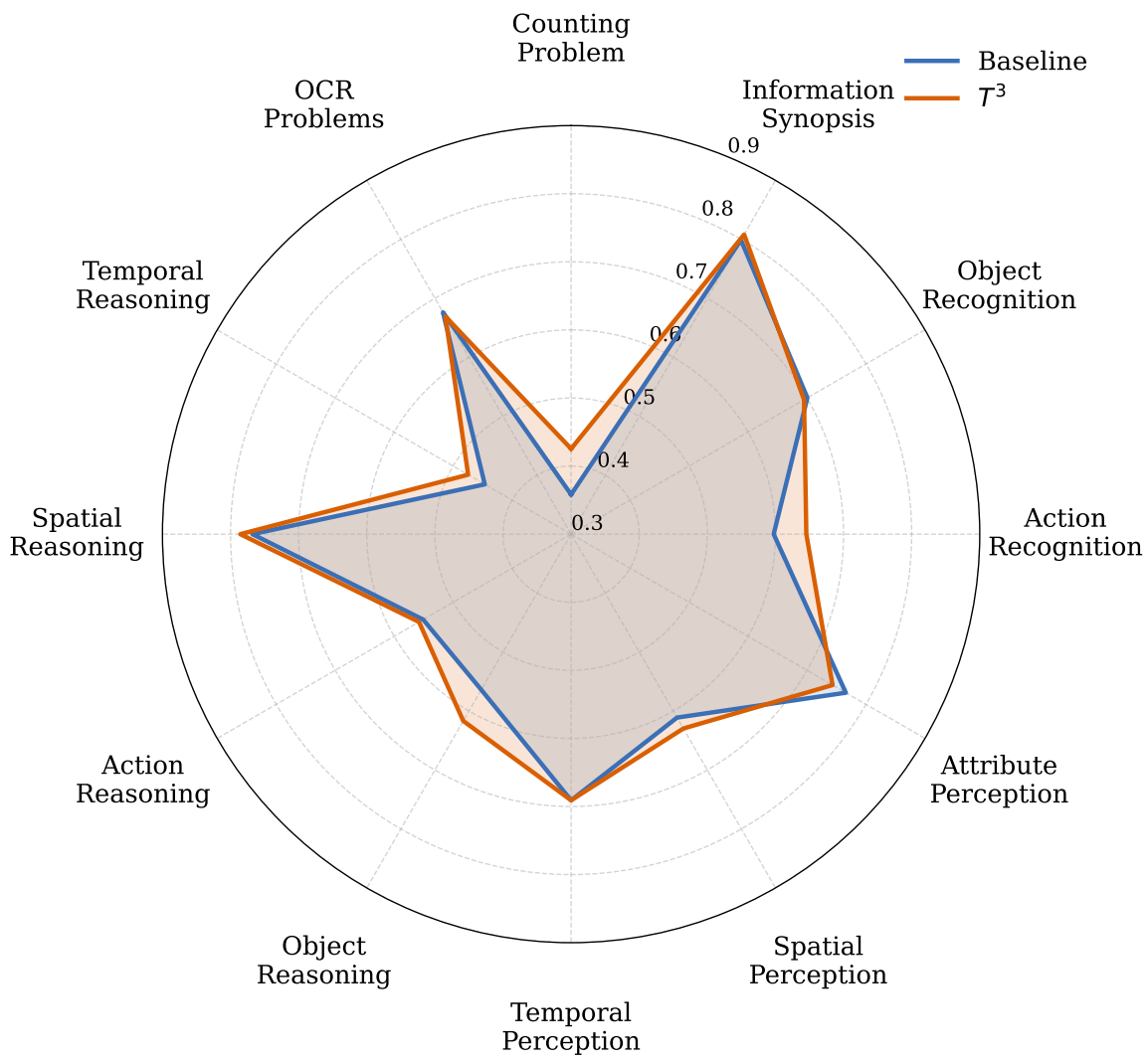

**Figure 4:** Task-type analysis on VideoMME using LLaVA-Video-7B as the backbone.

To better understand where $T^3$ brings improvements, we analyze the performance across different task types on VideoMME, as shown in Figure 4. Compared with the baseline, $T^3$ improves most reasoning-intensive categories, including Counting Problem, Object Reasoning, Action Recognition, Temporal Reasoning, and Spatial Reasoning. These gains suggest that the proposed temporal tree and multi-round reasoning are particularly helpful when the model needs to progressively localize relevant evidence and refine its focus toward fine-grained objects, actions, and temporal details.

We also observe improvements on Information Synopsis and Spatial Perception, indicating that the coarse-to-fine search process can preserve global context while retrieving useful local evidence. Meanwhile, $T^3$ performs comparably on Temporal Perception and shows slight drops on Object Recognition, Attribute Perception, and OCR Problems. This is reasonable because these categories are more perception-oriented and often depend on whether the key visual cue is already visible in the sampled frames, leaving less room for multi-round temporal exploration.

Overall, the results show that $T^3$ is most beneficial for reasoning-heavy and temporally grounded questions, which aligns with our motivation of using structured temporal search for long-video understanding.

## A.5 Runtime Analysis

Table 6 reports runtime normalized by video length (seconds per minute of video). The Qwen2.5-VL baseline does not use any offline preprocessing and spends $3.03$ sec/min-video for each question. $T^3$-Q adds a one-time offline temporal tree construction ($11.4$ sec/min-video) and then answers each question with an online cost of $4.34$ sec/min-video. This is a good trade-off when a video has multiple questions. For a video with $M$ questions, the total cost is

$$\begin{aligned} T_{\text{Qwen}}(M) &= 3.03M, \\ T_{\text{online}}(M) &= 7.66M, \\ T_{T^3\text{-ours}}(M) &= 11.4 + 4.34M. \end{aligned} \tag{1}$$

So the offline cost is paid once and can be shared by all questions, while the online cost is paid per question. Compared to the baseline, $T^3$-Q adds $+1.31$ sec/min-video per question (from $3.03$ to $4.34$). This extra time mainly comes from doing a few extra rounds to find missing evidence (rewrite the search sentence, retrieve more frames, and run the model again). In addition, we stop early when the model is confident. This keeps the runtime stable even for long videos. The online clustering variant highlights the benefit of reusable temporal tree construction. Performing temporal clustering for each question raises the online cost to 7.66 sec/min-video, compared with 4.34 sec/min-video for $T^3$-Q, indicating substantial repeated computation when the temporal structure is rebuilt per query. Overall, $T^3$-Q pays a one-time preprocessing cost and only adds a small extra time per question, while improving long-video understanding.

| Dataset | Question and Options | Round | Pred. / Conf. | Generated Search Sentence (GSS) |
|---|---|---|---|---|
| VideoMME | **Q:** Which of the following is NOT a reason for water scarcity mentioned in the video? GT: C<br>A. Climate change; B. High consumption by industrial production; C. Population growth leading to increased consumption of drinking water; D. High consumption by agricultural production. | 1 | A / 0.291 | The video does not mention any reason for water scarcity other than those listed in the options provided. |
| | | 2 | C / 0.746 | Identify scenes showing people discussing or presenting solutions related to water scarcity. |
| | | 3 | C / 0.820 | Identify scenes showing industrial water usage or pollution. |
| VideoMME | **Q:** What actions does the red-black parrot perform in this video? GT: A<br>A. Putting the ball in the basket; B. Throwing the ball; C. Eating the ball; D. Breaking the ball. | 1 | B / 0.492 | The red-black parrot is seen interacting with a person, possibly performing tricks or engaging in playful behavior. |
| | | 2 | A / 0.867 | The red-black parrot is shown in various settings, including a cage and an outdoor environment, highlighting its interactions and activities. |
| | | 3 | A / 0.967 | The red-black parrot is seen interacting with objects like a mirror and a basketball hoop, showcasing its playful behavior. |
| LVB | **Q:** In a room where all the furniture is made of solid wood, there is a person sitting on a chair holding an embroidery with blue fabric. Who is this person doing the embroidery? GT: D<br>A. A woman in a white dress; B. A woman in a blue dress; C. A woman in a black dress; D. A woman in a crimson dress. | 1 | A / 0.418 | A person is sitting on a chair in a room with solid wood furniture, holding an embroidery with blue fabric. |
| | | 2 | D / 0.680 | The person is engaged in embroidery work using a hoop and needle, surrounded by wooden decor. |
| | | 3 | D / 0.944 | The person is wearing a long-sleeved white shirt and a brown vest while working on the embroidery. |
| LVBench | **Q:** What's the number on the back of Tyson Gay? GT: D<br>A. 3218; B. 3712; C. 3718; D. 3217. | 1 | C / 0.478 | Tyson Gay's number is 3718. |
| | | 2 | D / 0.677 | Tyson Gay running in the 100m race at London 2012 Olympics. |
| | | 3 | D / 0.959 | Tyson Gay wearing jersey with number 3217 at London 2012 Olympics. |

**Table 5:** Examples of generated search sentence evolution across reasoning rounds. Rather than simply rephrasing the original question, the generated search sentence is updated according to the current evidence and intermediate prediction. Depending on the retrieved evidence, it may progressively zoom into finer visual details or shift to a different search target when the current hypothesis becomes insufficient.

| Method | Offline | Online |
|---|---|---|
| Qwen2.5-VL | – | 3.03 |
| $T^3$ w/ Online Clustering | – | 7.66 |
| $T^3$ (Ours) | 11.4 | 4.34 |

**Table 6: Runtime Analysis (sec/min-video).** We report time normalized by video length. Offline is computed once per video, while Online is per question.

### A.6 Video Length on Early Stopping.

Figure 5 illustrates how the termination behavior of $T^3$ varies with video length on the VideoMME dataset. For short videos, about 46% of samples terminate after the first round, indicating that the initial evidence is often sufficient for confident decision making. The remaining 54% enter later rounds for additional verification. As video length increases, early termination becomes less frequent: only 29% of medium-length videos stop after round 1, and for long videos, the majority (72%) require three rounds to reach a confident answer. This behavior reflects two common failure modes when reasoning over long videos. In some cases, the decisive cue is brief and easy to miss at a broad glance, so the model needs additional rounds to broaden and re-check alternative temporal regions. In other cases, the model has already located a plausible region but remains uncertain because the answer hinges on subtle visual details, which motivates further refinement of that region before committing to a prediction.

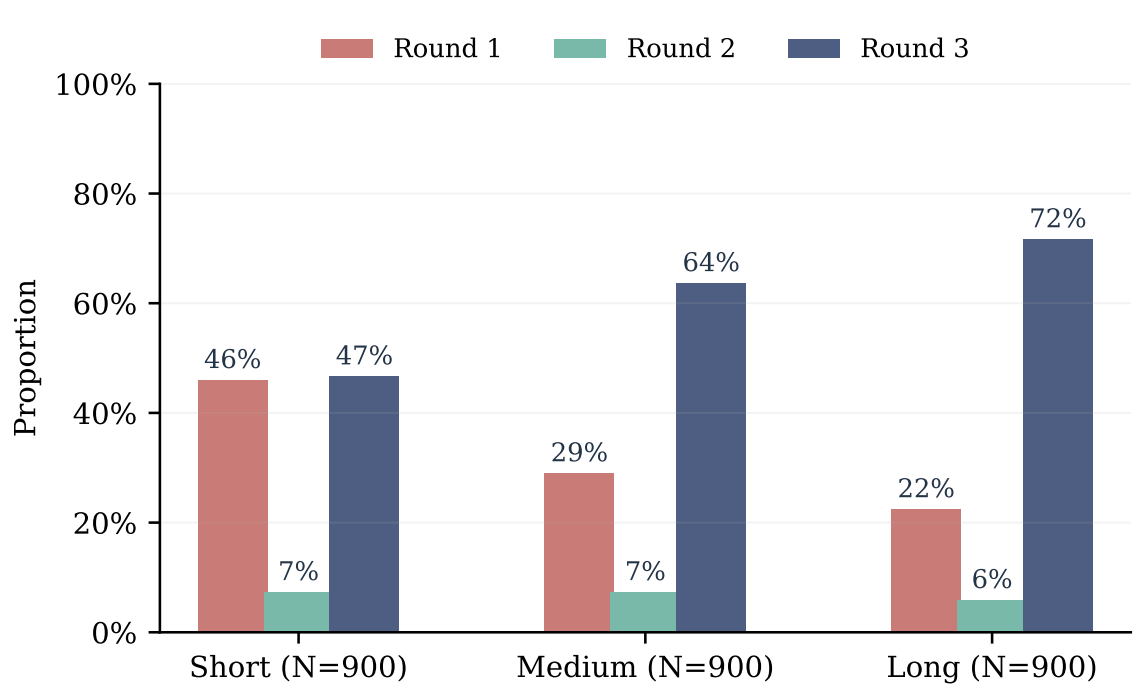

**Figure 5:** Distribution of early stopping rounds across short, medium and long on VideoMME dataset. N is the number of samples.

Interestingly, the proportion of samples terminating at round 2 is relatively small across all lengths. This reflects the role of the second round as an evidence consolidation stage: additional visual cues often weaken or revise an initially plausible but uncertain answer from round 1, yet may still be insufficient to reach high confidence. As a result, some samples proceed to a third round where decisive evidence is finally collected. This trend can also be seen in Figure 3.

### A.7 Choice of the Early Stopping Threshold.

To study the stopping criterion, we analyze the confidence–accuracy behavior of Qwen2.5-VL-7B under a 128 frame uniform sampling setting on validation subset of each dataset. We use uniform sampling for calibration as it provides a query-agnostic and method-independent confidence distribution. Specifically, we run the frozen MLLM on every sample in the validation or test set using uniformly sampled 128 frames on three datasets. For each sample, we record the predicted option and its confidence, defined as the next-token probability over the option letters. We then aggregate these records and sweep the confidence threshold $\tau$ to obtain accuracy–confidence curves.

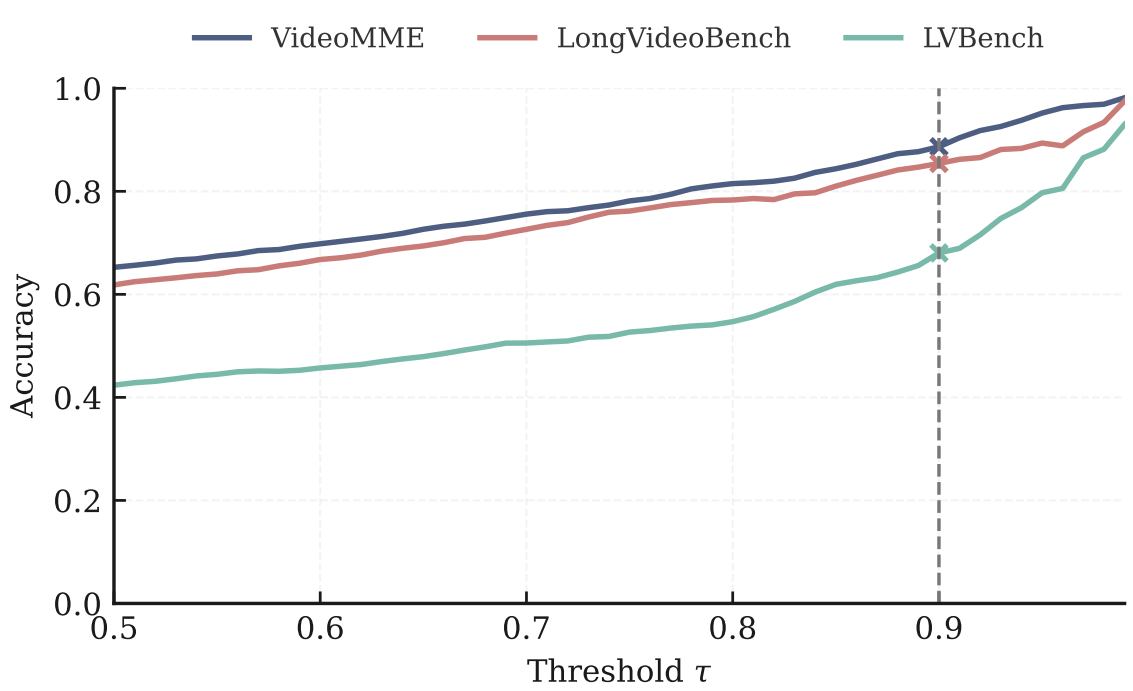

**Figure 6:** Accuracy–confidence curves on VideoMME, LongVideoBench, and LVBench. For each dataset, we sweep the early-stopping threshold $\tau$ over the next-token confidence For each $\tau$, we report the accuracy of predictions whose confidence is $\geq \tau$.

As shown in Figure 6, accuracy increases monotonically as $\tau$ becomes stricter, with the most pronounced gains occurring in the range $[0.6, 0.9]$. This suggests that low-confidence predictions are disproportionately likely to be incorrect, and filtering them improves reliability. Based on this dataset-level analysis, we set $\tau = 0.9$, which preserves most of the achievable accuracy while acting as a conservative trigger to avoid unnecessary expansions during reasoning.

| $\tau$ | Accuracy |
|---|---|
| 0.3 | 63.2 |
| 0.6 | 65.1 |
| 0.9 | **65.4** |

**Table 7:** Ablation study of the confidence threshold $\tau$ on VideoMME with $T^3$. Results are reported using Qwen2.5-VL-7B as the backbone.

After selecting the confidence threshold, we further conduct an ablation study on $\tau$ within $T^3$ on VideoMME, as shown in Table 7. When $\tau$ is small, e.g., $\tau = 0.3$, the model stops more aggressively and may terminate before collecting sufficient evidence, resulting in an accuracy of 63.2. Increasing $\tau$ to 0.6 improves the accuracy to 65.1, suggesting that additional reasoning is beneficial when the initial prediction is uncertain. The best result is achieved at $\tau = 0.9$, reaching 65.4 accuracy. This confirms that a conservative stopping threshold better matches the goal of $T^3$: the model stops only when the current evidence is reliable, and otherwise continues exploring the temporal tree to collect finer-grained evidence.

### A.8 Ablation Study for $K$ and $\rho$

Unless otherwise noted, all experiments are carried out using $T^3$-Q on the VideoMME dataset.

#### A.8.1 Effect of First-Layer Segments $K_1$.

| $K_1$ | Round 1 | Round 2 | Round 3 | Round 4 | Round 5 |
|---|---|---|---|---|---|
| 64 | 63.04 | **65.93** | 65.44 | 65.37 | 65.07 |
| 32 | 59.48 | 64.62 | **65.41** | **65.41** | 64.96 |
| 16 | 56.30 | 61.07 | 63.04 | 64.22 | **64.67** |

**Table 8: Effect of first-layer segments $K_1$ on VideoMME (Overall).** Values are overall accuracy per reasoning round; best per row in bold. Other settings fixed: $K_{>1}$=4, $\rho$=0.75.

Table 8 shows a clear coarse–to–fine trade-off. A finer first layer $K_1 = 64$ yields the best early performance, peaking at 65.93 in Round 2; with more, diverse first level temporal segments the model often finds sufficient evidence after just one refinement. As we reduce the first-layer granularity, the peak shifts to later rounds and drops slightly in value: $K_1$=32 tops out at 65.41 in Rounds 2–3, while $K_1$=16 requires more zoom-in and peaks at 64.67 in Round 5. This pattern is consistent with a fixed per-round frame budget: large $K_1$ improves coverage early, but can induce more subsampling and diminishing returns if we keep refining; small $K_1$ starts coarser, so additional rounds are needed to reach comparable evidence. In practice, $K_1$=64 is preferable when low-latency answers are desired (strong early rounds), whereas $K_1 = 32$ and $K_1 = 16$ trade a slower burn-in for slightly steadier later-round gains.

### A.8.2 Effect of Layer Branching $K_{>1}$.

| $K_1$=64 | | | | | |
|---|---|---|---|---|---|
| $K_{>1}$ | **Round 1** | **Round 2** | **Round 3** | **Round 4** | **Round 5** |
| 2 | 63.04 | 64.33 | 65.44 | **65.63** | 65.48 |
| 4 | 63.04 | **65.93** | 65.44 | 65.37 | 65.07 |
| 6 | 63.04 | 62.33 | **63.70** | 63.41 | 64.07 |
| $K_1$=16 | | | | | |
| $K_{>1}$ | **Round 1** | **Round 2** | **Round 3** | **Round 4** | **Round 5** |
| 2 | 56.30 | 59.07 | 61.44 | 62.52 | **63.48** |
| 4 | 56.30 | 61.07 | 63.04 | 64.22 | **64.67** |
| 6 | 56.30 | 62.33 | 63.70 | 63.41 | **64.07** |

**Table 9: Effect of deeper-layer branching ($K_{>1}$) on VideoMME (Overall).** Single table with two panels: top fixes $K_1$=64, bottom fixes $K_1$=16. Values are overall accuracy by reasoning round; best per row is in bold.

Table 9 compares the fan–out per node in deeper layers while fixing the first layer $K_1$=64 in the top panel; $K_1$=16 in the bottom). A consistent pattern emerges: a moderate branching factor, $K_{>1}$=4, performs best. With $K_1$=64, it peaks early at Round 2 65.93, indicating that a few children per segment provide enough local diversity to surface evidence quickly. With a coarser first layer $K_1$=16, $K_{>1}$=4 also wins, but the peak shifts later (Round 5, 64.67), as additional refinement is needed to recover details. In contrast, $K_{>1}$=2 under-explores (lower peaks), while $K_{>1}$=6 spreads the fixed per-round frame budget too thin, inducing heavier subsampling and flatter curves. Overall, $K_{>1}$=4 strikes the best balance between local exploration and token economy across both first-layer granularities.

### A.8.3 Effect of the selection ratio $\rho$.

Table 10 shows that a *moderate* ratio works best overall, and the optimal strictness depends on the first–layer granularity. With a finer first layer ($K_1$=64, top), $\rho$=0.75 reaches the highest value early at Round 2 65.93, while a looser gate $\rho$=0.70 requires more refinement and peaks at Round 4 ($\mathbf{0.6604}$). A stricter $\rho$=0.80 behaves similarly to Top–1 selection, under–explores, and only reaches 65.11 by Round 5. With a coarser first layer ($K_1$=16, bottom), the peak shifts later: $\rho$=0.70 peaks at Round 4 (65.15), slightly ahead of $\rho$=0.75 which peaks at Round 5 (64.67); $\rho$=0.80 is consistently weaker. In summary, $\rho$ acts as a soft Top–$k$ selector: too high removes complementary regions, too low expands too many segments and blurs the signal under a fixed frame budget, while $\rho \approx 0.75$ balances precision and coverage and gives strong early–round accuracy when the first layer is already fine.

| $K_1$=64 | | | | | |
|---|---|---|---|---|---|
| $\rho$ | **Round 1** | **Round 2** | **Round 3** | **Round 4** | **Round 5** |
| 0.70 | 63.04 | 64.48 | 65.15 | **66.04** | 65.70 |
| 0.75 | 63.04 | **65.93** | 65.44 | 65.37 | 65.07 |
| 0.80 | 63.04 | 64.74 | 64.78 | 64.85 | **65.11** |
| $K_1$=16 | | | | | |
| $\rho$ | **Round 1** | **Round 2** | **Round 3** | **Round 4** | **Round 5** |
| 0.70 | 56.30 | 61.41 | 63.52 | **65.15** | 64.33 |
| 0.75 | 56.30 | 61.07 | 63.04 | 64.22 | **64.67** |
| 0.80 | 56.29 | 61.04 | 63.74 | **64.07** | 63.63 |

**Table 10: Effect of the selection ratio $\rho$ on VideoMME (Overall).** Two panels fix the first-layer partition at $K_1$=64 (top) and $K_1$=16 (bottom). Values are overall accuracy per reasoning round; best per row in bold.

| **Error type** | **#Cases** | **Ratio** |
|---|---|---|
| `found_incorrect` | 465 | 57.1% |
| `notfound_incorrect` | 350 | 42.9% |

**Table 11:** Breakdown of incorrect predictions on LVBench.

## B Failure Cases Analysis

We analyze errors on LVBench using its provided evidence timestamps, as in Table 11. Among 815 incorrect predictions, we observe two dominant failure modes: found_incorrect (465, 57.1%) and notfound_incorrect (350, 42.9%). found_incorrect denotes cases where the model retrieved relevant segments but still answered incorrectly. Qualitatively, these errors often stem from (i) hallucinations or (ii) insufficiently discriminative evidence within the retrieved context. notfound_incorrect corresponds to failures where the required evidence is never retrieved. This mode is strongly associated with sparse evidence: LVBench videos are long (average 4,101s), yet the decisive evidence may occupy only a few frames around the ref_time. Such extreme temporal sparsity makes the signal easy to miss even with multi-round reasoning.

## C Prompts

Figure 7 shows the QA prompt format used in our experiments.

**QA Prompt**

```
Select the best answer to the following multiple-choice question based on the video. Respond with only the letter of the correct option.
Question: {q}
A. {o_1}
B. {o_2}
C. {o_3}
D. {o_4}
Answer (one of: A, B, C, D):
```

**Figure 7:** Prompt template for video multiple-choice QA.

Figure 8 shows the prompt format used to generate search sentences for iterative frame retrieval.

**Generate Search Sentences Prompt**

```
You will see several video frames, the original question, and the previous search sentence. The model's answer is not yet confident.
Propose 1 alternative short declarative search statements that could help retrieve additional relevant frames to increase confidence. Each statement should name concrete, visible cues (objects, actions, scenes, coarse temporal hints). Do not ask a question. Do not include answer options or an answer. Output exactly 1 lines, one statement per line, no numbering and no quotes.
Original question: {q}
Previous search sentence: {s_prev}
Now output the 1 statements:
```

**Figure 8:** Prompt template for generating search sentences.

## D Pseudo Code for Temporal Clustering

The pseudo-code for our temporal clustering is shown in Algorithm 1.

### D.1 Overall Purpose of the Algorithm

The algorithm performs temporal segmentation on a video sequence. Its goal is to divide the sequence into K contiguous, non-overlapping subsegments, each satisfying a minimum-length constraint, in a way that minimizes the total within-segment sum of squared errors (SSE). Once the optimal segmentation is obtained, the algorithm computes the mean feature vector for each subsegment, which is then used as the cluster center for that subsegment.

**Algorithm 1** Temporal Clustering for a Single Segment

**Input:** $X = \{x_1, \ldots, x_T\}$, $K$, $L_{\min}$
**Output:** $\mathcal{S} = \{[s_1, e_1], \ldots, [s_K, e_K]\}$
**Center:** $C = \{\mu_k\}_{k=1}^{K}$
1: **Prefix sums:** $S_x[0] \leftarrow 0$, $S_{xx}[0] \leftarrow 0$; for $t$=1..$T$: $S_x[t] \leftarrow S_x[t-1]+x_t$, $S_{xx}[t] \leftarrow S_{xx}[t-1]+\langle x_t, x_t \rangle$
2: **Interval cost:** $\mathrm{SSE}(i,j) = S_{xx}[j] - S_{xx}[i-1] - \frac{\|S_x[j]-S_x[i-1]\|_2^2}{j-i+1}$
3: **Init:** $dp[0][0] \leftarrow 0$; otherwise $+\infty$; $prv \leftarrow -1$
4: **for** $k = 1..K$ **do**
5:   **for** $t = kL_{\min}..T$ **do**
6:     $best \leftarrow +\infty$, $best_p \leftarrow -1$; $p_{\min} \leftarrow (k-1)L_{\min}$, $p_{\max} \leftarrow t-L_{\min}$
7:     **for** $p = p_{\min}..p_{\max}$ **do**
8:       $cost \leftarrow dp[k-1][p] + \mathrm{SSE}(p+1, t)$
9:       **if** $cost < best$ **then**
10:         $best \leftarrow cost$; $best_p \leftarrow p$
11:       **end if**
12:     **end for**
13:     $dp[k][t] \leftarrow best$; $prv[k][t] \leftarrow best_p$
14:   **end for**
15: **end for**
16: **Backtrack:** $t \leftarrow T$; $\mathcal{S} \leftarrow \emptyset$
17: **for** $k = K..1$ **do**
18:   $p \leftarrow prv[k][t]$; prepend $[p+1, t]$ to $\mathcal{S}$; $t \leftarrow p$
19: **end for**
20: **Compute centers:** for $k = 1..K$, $\mu_k \leftarrow \frac{1}{e_k - s_k + 1} \sum_{t=s_k}^{e_k} x_t$
21: **return** $(\mathcal{S}, C)$

### D.2 Inputs and Outputs

**Inputs** The algorithm takes the following inputs:

- **Sequence** $X = \{x_1, \ldots, x_T\}$: An ordered collection of frame features.
- **Number of segments** $K$: Specifies the number of contiguous segments into which the sequence will be divided.
- **Minimum segment length** $L_{\min}$: Ensures that each segment contains at least $L_{\min}$ consecutive video frames.

**Outputs** The algorithm produces two outputs:

- **Segmentation** $\mathcal{S}$:

$$\mathcal{S} = \{[s_1, e_1], \ldots, [s_K, e_K]\}.$$

Each pair $[s_k, e_k]$ denotes the start index $s_k$ and end index $e_k$ of segment $k$. The segments are contiguous, non-overlapping, satisfy the minimum-length constraint, and together form the optimal partition minimizing the total SSE.

- **Segment centers** $C = \{\mu_k\}_{k=1}^{K}$: For each segment $[s_k, e_k]$, the center $\mu_k$ is computed as the mean of all frame features within that interval, serving as the cluster center features for that segment.

# E Pseudo Code for Stopping Criterion

The pseudo-code for extracting option probabilities from the model's generation output is shown in Algorithm 2.

## E.1 Overall Purpose of the Algorithm

The algorithm computes normalized probabilities over four multiple-choice options $\{A, B, C, D\}$ from a generation output. It reads the first-step logits (i.e., the scores of the first generated token), maps each option letter to its tokenizer ID, gathers the logits for the available option tokens, and applies a softmax to obtain a probability distribution over the available options. We manually verify all outputs and confirm that the first generated token is always one of the valid option letters, i.e., $\{A, B, C, D\}$.

---

**Algorithm 2** Extract Option Probabilities from Generation Output

---

**Input:** `output`, `tokenizer`
**Output:** `probs` $= \{$`A` : $\rho_A$, `B` : $\rho_B$, `C` : $\rho_C$, `D` : $\rho_D\}$

1: **Check scores:** if `output` has no attribute `scores` or $|$`output.scores`$| = 0$, return $\{$`A` : `None`, `B` : `None`, `C` : `None`, `D` : `None`$\}$
2: **Read logits:** $\ell \leftarrow$ `float32(output.scores[0])` $\{\ell$ is the first-step logit tensor$\}$
3: **Token id mapping:** for each $k \in \{$`A`, `B`, `C`, `D`$\}$:
4:     `ids` $\leftarrow$ `tokenizer.encode`($k$, `add_special_tokens` = `False`)
5:     `id`$[k] \leftarrow$ `ids`$[0]$ if $|$`ids`$| > 0$ else `None`
6: **Gather option logits:** `picked` $\leftarrow [\,]$, `keys` $\leftarrow [\,]$
7: **for** $k \in \{$`A`, `B`, `C`, `D`$\}$ **do**
8:    `tid` $\leftarrow$ `id`$[k]$
9:    **if** `tid` $\neq$ `None` **then**
10:       append $\ell[0,$ `tid`$]$ to `picked`; append $k$ to `keys`
11:    **end if**
12: **end for**
13: **if** $|$`picked`$| = 0$ **then**
14:    return $\{$`A` : `None`, `B` : `None`, `C` : `None`, `D` : `None`$\}$
15: **end if**
16: **Normalize:** $\pi \leftarrow$ `softmax(stack(picked))` $\{\pi$ is over available options only$\}$
17: **Fill output:** initialize `probs`$[k] \leftarrow$ `None` for all $k \in \{$`A`, `B`, `C`, `D`$\}$
18: **for** $(k, p) \in$ `zip(keys`, $\pi)$ **do**
19:    `probs`$[k] \leftarrow p$
20: **end for**
21: **return** `probs`

---

## E.2 Inputs and Outputs

**Inputs** The algorithm takes the following inputs:

- **Generation output `output`**: The model's generation result object. The algorithm expects `output.scores` to exist and contain the per-step logits (at least for the first generated token).

- **Tokenizer `tokenizer`**: A tokenizer providing `encode`$(\cdot)$ for mapping the option letters $\{$`A`, `B`, `C`, `D`$\}$ to token IDs.

**Outputs** The algorithm produces the following output:

- **Option probability dictionary `probs`**:

$$\texttt{probs} = \{\texttt{A} : \rho_A,\ \texttt{B} : \rho_B,\ \texttt{C} : \rho_C,\ \texttt{D} : \rho_D\}.$$

  For each option $k \in \{\texttt{A}, \texttt{B}, \texttt{C}, \texttt{D}\}$, $\rho_k$ is either a normalized probability (softmax over the available option logits) or `None` if scores are missing or the corresponding token ID cannot be obtained. The output distribution is normalized over the subset of options whose token IDs are valid.